\documentclass[letterpaper]{article}
\usepackage{aaai18}
\usepackage{times}
\usepackage{helvet}
\usepackage{courier}
\usepackage{url}
\usepackage{graphicx}
\usepackage{booktabs}
\usepackage{tabularx}
\usepackage{multirow}
\usepackage{amssymb}
\usepackage{amsmath}
\usepackage{bm}
\usepackage{pgfplots}
\usepackage{subfig}
\usepackage{threeparttable}
\usepackage{wrapfig}
\usepackage{soul}
\usetikzlibrary{patterns}
\begin{document}

\title{Medical Exam Question Answering with Large-scale Reading Comprehension}
\author{Xiao Zhang\textsuperscript{1}, 
Ji Wu\textsuperscript{1}, 
Zhiyang He\textsuperscript{1}, 
Xien Liu\textsuperscript{2},
Ying Su\textsuperscript{2}\\
\textsuperscript{1}Department of Electronic Engineering, Tsinghua University\\
zhang-xi15@mails.tsinghua.edu.cn, \{wuji\_ee, zyhe\_ts\}@mail.tsinghua.edu.cn\\
\textsuperscript{2}Tsinghua-iFlytek Joint Laboratory, Tsinghua University\\
\{xeliu, yingsu2\}@iflytek.com}

\maketitle
\begin{abstract}
Reading and understanding text is one important component in computer aided diagnosis in clinical medicine, also being a major research problem in the field of NLP. 
 In this work, we introduce a question-answering task called MedQA to study answering questions in clinical medicine using knowledge in a large-scale document collection. The aim of MedQA is to answer real-world questions with large-scale reading comprehension. We propose our solution SeaReader---a modular end-to-end reading comprehension model based on LSTM networks and dual-path attention architecture. The novel dual-path attention models information flow from two perspectives and has the ability to simultaneously read individual documents and integrate information across multiple documents. In experiments our SeaReader achieved a large increase in accuracy on MedQA over competing models.  Additionally, we develop a series of novel techniques to demonstrate the interpretation of the question answering process in SeaReader.
\end{abstract}

\section{Introduction}
Natural language understanding is a crucial component in computer aided diagnosis in clinical medicine. Ideally, a computer model could read text to communicate with human doctors and utilize knowledge in text materials. Recent advances in deep-learning-inspired approaches have taken the state-of-the-art of various NLP tasks to a new level. However, directly reading and understanding text is still a challenging problem, especially in complex real-world scenarios.

Reading comprehension is a task designed for reading text and answering questions about it. An understanding of natural language and basic reasoning is required to tackle the task. The task of reading comprehension has been gaining rapid progress with the proposal of various datasets such as SQuAD \cite{2016arXiv160605250R} and many successful models.

Real-world scenarios for reading comprehension are usually much more complex. Unlike in most datasets, one does not have paragraphs of text already labeled as containing the answer to the question. Rather, one needs to find and extract relevant information from possibly large-scale text materials. 
Previous work \cite{miller2016key} and \cite{chen2017reading} reads from Wikipedia dump to answer questions created from knowledge base entries. Another major challenge in real-world scenarios is that questions are often harder and more versatile, and the answer is less likely to be directly found in text.

We propose the MedQA, our reading comprehension task on clinical medicine aiming at simulating a real-world scenario. Computers need to read from various sources in aided diagnosis, such as patients' medical records and examination reports. They also need to exploit knowledge found in text materials such as textbooks and research articles. We assembled a large collection of text materials in MedQA as a source of information, to learn to read large-scale text. 

Questions are taken from medical certification exams, where human doctors are evaluated on their professional knowledge and ability to make diagnosis. Questions in these exams are versatile and generally requires an understanding of related medical concepts to answer. A machine learning model must learn to find relevant information from the document collection, reason over them and make decisions about the answer.

We then propose our solution SeaReader: a large-scale reading comprehension model based on LSTM network and dual-path attention architecture. The model addresses challenges of the MedQA task from two main aspects: 1) \textbf{Leveraging information in large-scale text}: we propose a dual-path attention architecture which uses two separate attention paths to extract relevant information from individual documents as well as compare and extract information across multiple documents. Extracted information is reasoned over and integrated together to determine the answer. 2) \textbf{End-to-end training on weak labels}: although we only have labels with very little information, we still managed to train our model end-to-end. Our SeaReader features a modular design, with a clear interpretation of different levels of language understanding. We also propose a series of novel techniques that we call Interpretable Reasoning. This increases the interpretability of complex NLP models like our SeaReader. 

\newcolumntype{s}{>{\hsize=.8\hsize}X}
\begin{table*}[t]
  \scriptsize
  \centering
  \caption{Questions by category}
  \label{tab:q}
  \begin{tabularx}{\textwidth}{cscX}
    \toprule
    Category & Description & Ratio & Example \\
    \midrule
    A1 & \textbf{Single statement best choice}: single statement questions, 1 best answer, 4 incorrect or partially correct answers & 36.8\% & \multirow{2}{*}{\parbox[t][][t]{1.2\linewidth}{\textbf{Question}: The pathology in the pancreas of patients with type 1 diabetes is:
\\ \textbf{Candidate Answers}: a. Islet cell hyperplasia
b. Islet cell necrosis
c. interstitial calcification
d. Interstitial fibrosis
e. Islet cell vacuolar degeneration}}
    \\
    B1 & \textbf{Best compatible choice}: similar to A1, with a group of candidate answers shared in multiple questions & 11.2\% & \\
    A2 & \textbf{Case summary best choice}: questions accompanied by a brief summary of patient's medical record, 1 best choice among 5 candidate answers & 35.3\% & 
    \multirow{2}{*}{\parbox[t][][t]{1.2\linewidth}{\textbf{Question}: Male, 24 years old. Frontal edema, hematuria with cough, sputum with blood for 1 week. bp 160/100 mmhg, urinary protein (++), rbc 30/ hp, ... (omitted text) Ultrasound show kidney size increase. At present the most critical treatment is:
\\ \textbf{Candidate Answers}: a. hemodialysis
b. prednisone
c. plasma exchange
d. gamma globulin
e. prednisone combined with cyclophosphamide}
}\\
    A3/A4 & \textbf{Case group best choice}: similar to A2, with information shared among multiple questions & 16.7\% &\\
    \bottomrule
  \end{tabularx}
\end{table*}

\section{Related Work}
Our work is closely related to question answering and reading comprehension in the field of NLP.

\textbf{Reading Comprehension Tasks} Datasets are crucial for developing data-driven reading comprehension models. The Text REtrieval Conference (TREC) \cite{voorhees2000overview} evaluation tasks and the MCTest \cite{richardson2013mctest} are successful early efforts at creating datasets for answering questions based text documents. Large-scale datasets have become more popular in the era of deep learning. The CNN/DailyMail dataset \cite{hermann2015teaching} creates a combination of over 1 million questions on news articles by removing words in summary points. The Children's Book dataset takes a similar approach and use excerpts from children's books as reading materials. The SQuAD \cite{2016arXiv160605250R} dataset features 100K Wikipedia articles and human created questions. Answers are spans of text instead of single words in the reading passage, which creates more challenge in answer selection. 

Some larger-scale question answering datasets aim to create a more real-world situation. For example, the WikiMovies benchmark \cite{miller2016key} asks questions about movies, where Wikipedia articles and knowledge-base (OMDb) can be leveraged to answer the question. The webQA dataset \cite{li2016dataset} collects 42,000 questions about daily life asked by users on QA site. The MS MARCO dataset \cite{nguyen2016ms} takes query logs of the users of Bing search engine as questions and uses passages extracted from web documents as reading material.

\textbf{Neural Network Models for Reading Comprehension} Numerous models have been proposed for this kind of tasks. Earlier attempts directly use LSTM to process text and uses attention to fetch information from the passage representation (Attentive Reader and the Impatient Reader, \citeauthor{hermann2015teaching}\citeyear{hermann2015teaching}). Later models largely rely on LSTMs and attention as the main building blocks. The Attention Sum Reader \cite{kadlec2016text} predicts candidate word by summing attention weights on occurrences of the same word. The Gated Attention Reader \cite{dhingra2016gated} uses gating units to combine attention over the query into the passage. It also performs multi-pass readings over the document. 

The Match-LSTM \cite{dhingra2016gated} uses separate LSTM layers to preprocess text input and predict answer. A span is selected from passage by predicting the start and end positions of the answer. Epireader \cite{trischler2016natural} differs from the above approaches by factoring question-answering into a two-stage process of candidate extraction and reasoning. Convolutional neural networks are used to generate sentence encodings in the reasoning of hypotheses.

Several state-of-the-art models make further improvements and are competitive on the challenging SQuAD dataset. BiDAF \cite{seo2016bidirectional} incorporates attention from passage to query as well as from query to passage. R-NET \cite{rnet} adds a passage-to-passage self attention layer on top of question-passage attention. The AoA Reader \cite{cui2016attention} shares the idea of utilizing both passage to query and query to passage attention, multiplying them together to make final prediction.

However, these models largely restrict themselves to selecting word(s) from the given passage, by modeling at the resolution of words. This makes them less straightforward to be applied to other tasks involving reading and understanding. They also rely on pre-selected relevant documents---a requirement not easily met in real-world settings.

\section{The MedQA Task}
The MedQA task answers questions in the field of clinical medicine. Rather than relying on memory and experience as human doctors do, a machine learning model may make use of a large collection of text materials to help find supporting information and reason about the answer.

\subsection{Task Definition}
The task is defined by its three components:
\begin{itemize}
\item Question: questions in text, possibly with a short description of the situation (of the patient).
\item Candidate answers: multiple candidate answers are given for each question, of which one should be chosen as the most appropriate.
\item Document collection: a collection of text material extracted from a variety of sources organized into paragraphs and annotated with metadata.
\end{itemize}
The goal is to determine the best candidate answer to the question with access to the documents.

\subsection{Data}
\subsubsection{Question and Answers} 
The goal of MedQA is to ask questions relevant in real-world clinical medicine that requires the ability of medical professionals to answer. We used the National Medical Licensing Examination\footnote{http://www.nmec.org.cn/EnglishEdition.html} in China as a source of questions. The exam is a comprehensive evaluation of professional skills of medical practitioners. Medical students are required to pass the exam to be certified as a physician in China. 

The General Written Test part of the exam consists of 600 multiple choice problems over 5 categories (see Table \ref{tab:q}). We collected over 270,000 test problems from the internet and published materials such as exercise books. The problems do not have to have appeared in past exams. We filtered out any incomplete or duplicate problems. The statistics of the final problem set are shown in Table \ref{tab:data_stat}. 

For training/test split, we created a test set as similar as possible to past exam problems, to approximate evaluation in real exams. A small subset of problems was chosen based on the source and context in which they appear.  This might indicate their possible appearance in past exams, or they might be closely related to past exam problems. These problems are further split into valid/test sets. The remaining problems that are similar to any one problem in the valid/test set are removed to ensure that one cannot solve test problems by merely remembering problems in the training set. The similarity of two problems is measured by comparing Levenshtein distance \cite{levenshtein1966binary} of questions with a threshold ratio of 0.9.
\begin{table}[h]
  \small
  \centering
  \caption{Data statistics}
  \label{tab:data_stat}
  \begin{tabular}{cccc}
    \toprule
     & training & valid & test  \\
    \midrule
    Number of problems & 222,323 & 6,446 & 6,405\\
    \bottomrule
    
  \end{tabular}
  \begin{tabular}{lc}
    \toprule
    \textit{Problems} &\\
    Average length of question (words) & 27.4 \\
    Average length of answer (words) & 4.2\\
    Candidate answers per problem & 5\\
    \midrule
    \textit{Documents} &\\
    Number of documents & 243,712\\
    Average length of document (words) & 43.2\\
    Average number of tags in document metadata & 3.8 \\
    \bottomrule
  \end{tabular}
\end{table}

\subsubsection{Documents}
Human knowledge in almost every modern profession is extensively encoded in text. Medical students obtain their knowledge from a volume of textbooks during years of training. A machine learning model, however, can easily access a large collection of text materials at their disposal. We prepared text materials from a total of 32 publications including textbooks, reference books, guidebooks, books for text preparation, etc. These books cover a wide range of topics in clinical medicine.

\begin{figure}[h]
\begin{minipage}{.50\columnwidth}
  \centering
  \includegraphics[width=\linewidth]{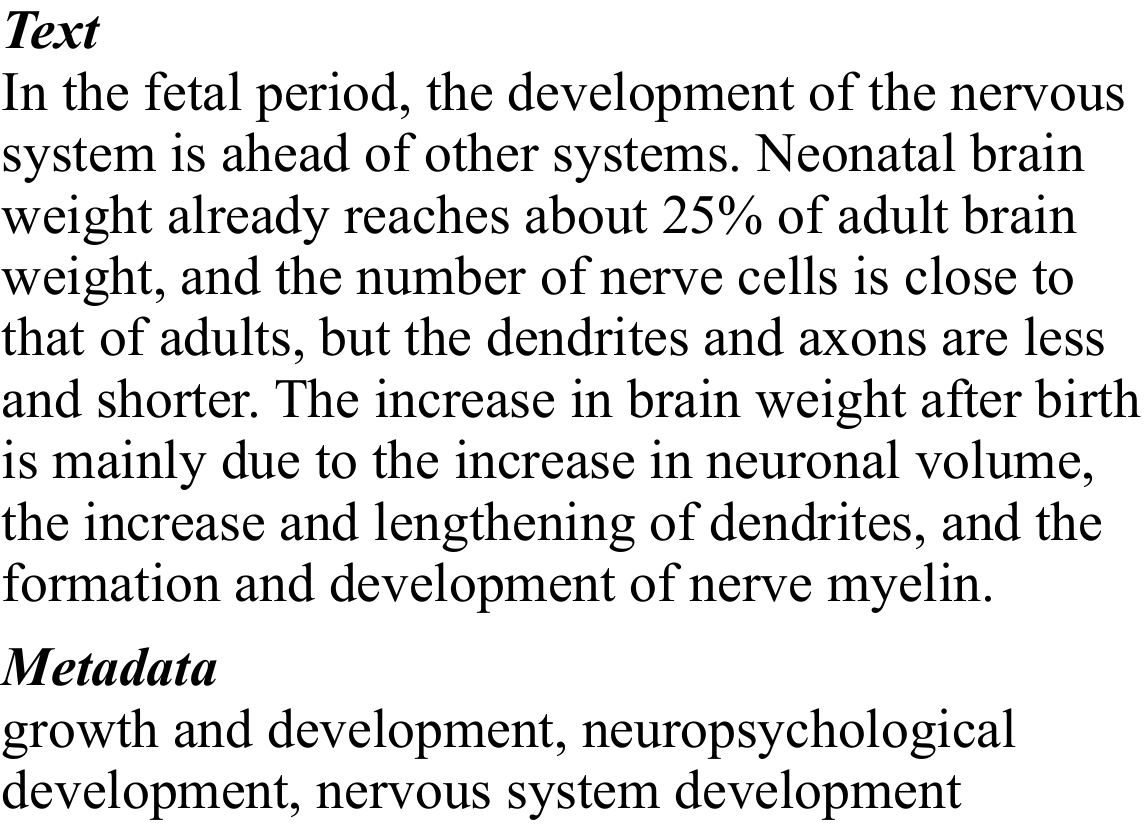}
  \caption{An example document}
  \label{fig:doc_eg}
\end{minipage}
\begin{minipage}{.1\columnwidth}
\end{minipage}
\begin{minipage}{.46\columnwidth}
  \centering
  \pgfplotsset{width=1.16*\columnwidth,compat=1.5,
every axis/.append style={label style={font=\tiny},
                    tick label style={font=\fontsize{3}{4}\selectfont}}
}
\begin{tikzpicture} \begin{axis}[
    bar width=1.6pt,
    ymin=0,
    ymax=30000,
	xlabel={Number of words},
    ylabel={Number of documents},
	xlabel shift=-4pt,
	xticklabel shift=-2pt,
	ylabel shift=-4pt,
	yticklabel shift=-2pt,
]
\addplot[ybar,fill=brown] coordinates {(0,12810) (5,24671) (10,28418) (15,26104) (20,21707) (25,16749) (30,13980) (35,11818) (40,10554) (45,8670) (50,7554) (55,6449) (60,5744) (65,5037) (70,4418) (75,3889) (80,3587) (85,3148) (90,2766) (95,2588) (100,2209) (105,2034) (110,1808) (115,1625) (120,1447) (125,1312) (130,1121) (135,1023) (140,945) (145,905) (150,817) (155,663) (160,672) (165,588) (170,531) (175,472) (180,430) (185,417) (190,351) (195,321)};

\end{axis}
\end{tikzpicture}
\caption{Length distribution of documents}
  \label{fig:doc_dist}
\end{minipage}
\end{figure}

Text material is extracted from these books, divided by paragraph, and annotated with metadata. Metadata include contextual information like tags, book titles and chapter titles. An example document is shown in Figure \ref{fig:doc_eg}.  Basic statistics of documents after pre-processing are given in Table \ref{tab:data_stat} and Figure \ref{fig:doc_dist}.

\subsection{Task Analysis}
The MedQA task poses unique challenges for language understanding, especially compared with existing reading comprehension datasets. We identify several major challenges of MedQA as below:
\begin{itemize}
\item \textbf{Professional knowledge} Unlike most reading comprehension tasks---where question answering rely largely on commonsense reasoning and basic understanding of language---MedQA asks questions in a field of sophistication, that usually requires a thorough understanding of the field for human to answer.
\item \textbf{Diversity of questions} The field of clinical medicine is diverse enough for questions to be asked about a wide range of topics. Questions can also be asked in various facets, for example: given \textit{a description of a patient's condition}, a question might ask for \textit{the most probable diagnosis / the most appropriate treatment / the examination needed / the mechanism of certain condition}, etc.
\item \textbf{Determining the best answer} Choosing the best answer means that the unselected answers are \textit{not} necessarily incorrect. The model must learn to evaluate individual answers and learn to make comparison.
\item \textbf{Reading large-scale text} Retrieving relevant information from large-scale text is more challenging than reading a short piece of text.  Furthermore, in MedQA passages from textbooks often do not directly give answers to questions, especially for \textit{case} problems. One must discern relevant information scattered in passages, and determine the relevance of each piece of text. 
\item \textbf{Reasoning over facts} Reasoning is often required to answer the questions. This includes natural language reasoning where one recognizes lexical or syntactic variations and reasoning over facts to decide whether given document(s) support an answer to the question. Below is an example of a problem requiring reasoning over multiple facts, like those in \cite{weston2015towards}:

\small{
\textit{Question}: The most effective treatment for a 70 years old patient with Parkinson's  disease is:\\
\textit{Document 1}: Commonly used drugs are anticholinergic drugs phenanthrene, amantadine, levodopa and compound levodopa\\
\textit{Document 2}: Phenanthrene is mainly suitable for those with obvious tremble, but is more used in young patients, elderly patients should be used with caution\\
} 
\end{itemize}

\section{The SeaReader Framework}
\begin{figure}[h]
  \centering
    \includegraphics[width=\columnwidth]{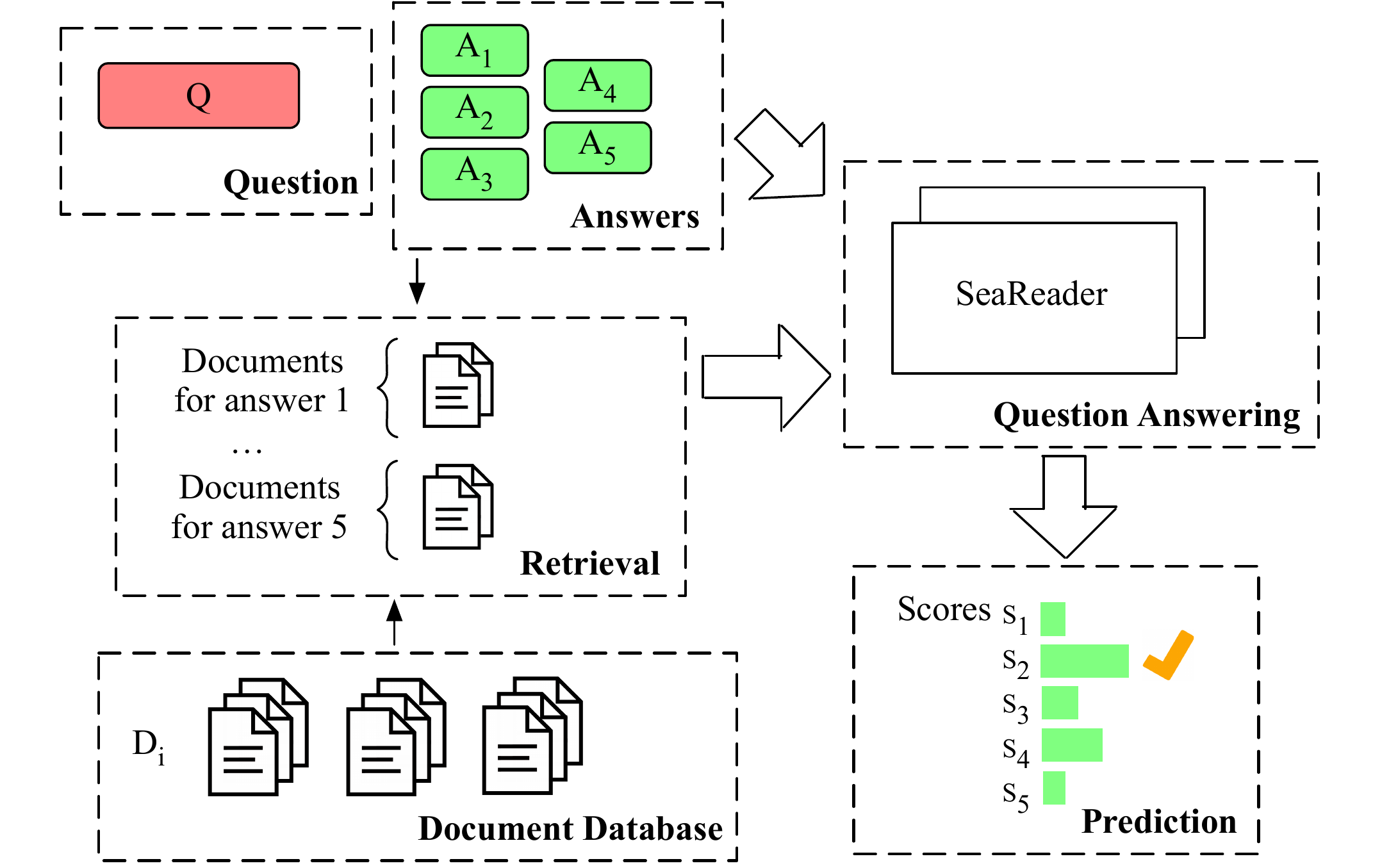}
  \caption{System overview}
  \label{fig:overview}
\end{figure}

Our proposed solution to the MedQA task includes a document retrieval system and a neural network based question answering model. Text retrieval is used to narrow down possibly related documents that are then fed into the SeaReader where the reasoning and decision-making occur. See Figure \ref{fig:overview} for an overview of the framework.

\subsection{Document Retrieval}
Given a problem, we select a subset of documents that are likely to be relevant using a text retrieval system. The text retrieval system is built upon Apache Lucene, using inverted index lookup (metadata included in documents) followed by BM25 ranking. For each problem, we perform retrieval for each answer individually by pairing it with the question. We take the intersection of the retrieved document for question and answer and keep the top-N documents. To account for the different nature of various text sources, we try to select an equal number of documents from each type of books. We found that this promotes diversity in the selected documents and improves the final performance of question answering.

\subsection{The SeaReader Model}

\begin{figure*}[t]
  \centering
    \includegraphics[width=\textwidth]{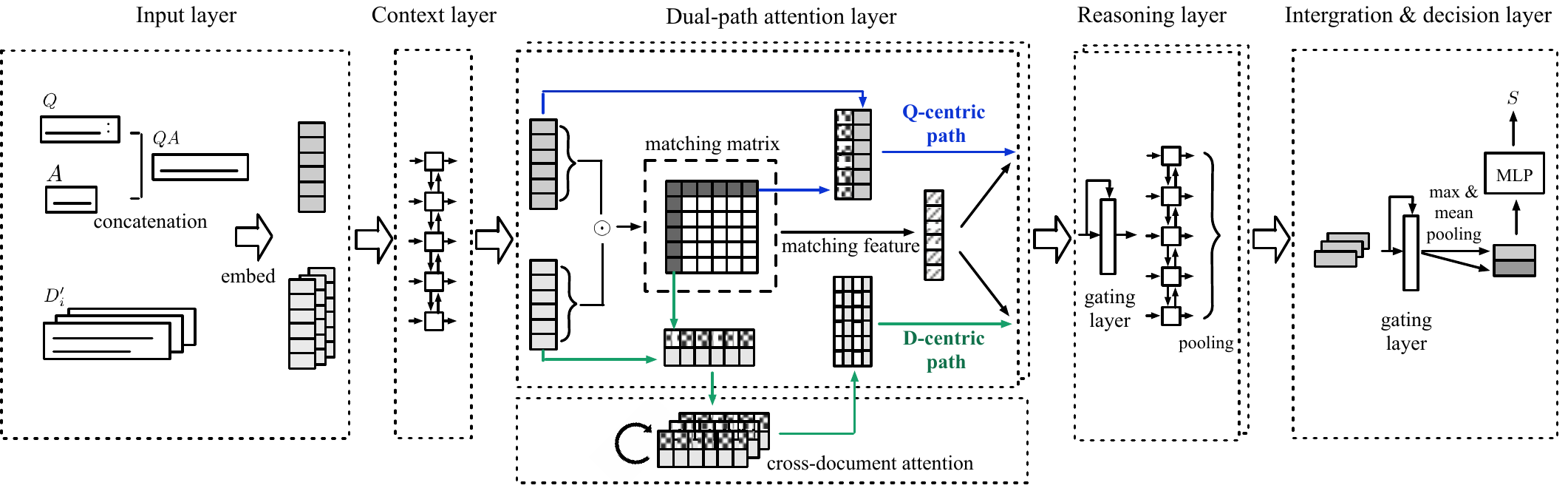}
  \caption{The SeaReader model}
  \label{model}
\end{figure*}

The Support Evidence Analysis Reader (SeaReader) model is designed to find relevant evidence from documents given a question and a candidate answer  and perform analysis and reasoning to determine the correctness of the answer. SeaReader uses LSTM networks to model context representations of text. Attention is used extensively in the dual-path attention architecture to model information flow between question and documents, and across multiple documents.
Information from multiple documents is fused together to make the final prediction. SeaReader also has a modular design that facilitates interpretation of the reasoning process, as discussed in the next section.
\subsubsection{Input layer} The input to SeaReader is a question-answer pair $(Q,A)$ (the answer being one of the candidate answers given in the problem) and the set of documents $\{D_0, D_1, ...,D_N\}$ returned by the retrieval system. The answer is appended to the question to form a statement $S$. After word-embedding lookup, we have tensors representing statement $S \in \mathbb{R}^{L_Q\times d}$ and documents $D \in \mathbb{R}^{N\times L_D\times d}$. $L_Q$ and $L_D$ are the maximum length of $S$ and $D_i$ respectively, and $d$ is the dimension of word-embedding.
\subsubsection{Context layer} The word-level representations are then processed  by a bi-directional LSTM layer to model contextual representations. Leaving out this layer, replacing it with a convolutional layer or GRU \cite{cho2014properties} all results in performance drop, indicating the importance of long-range context in representing semantics.
\subsubsection{Dual-path attention layer} We would like the model to learn to identify and extract relevant information from long documents. The dual-path attention architecture allows for extracting information from two perspectives: in the question-centric (Q-centric) path, information related to the question is extracted from documents and is aligned with the question. Information from each document is processed individually. In the document-centric (D-centric) path, we take the perspective of documents: information from the question is integrated to documents, then information from other documents is compared and integrated. Interaction of supporting facts in multiple documents is captured in this path.

Similar to several state-of-the-art works \cite{cui2016attention,seo2016bidirectional,xiong2016dynamic}, we start by computing a matching matrix as the dot-product of the context embeddings of the question and every document:
\begin{equation}
\small
	M_n(i,j) = S(i) \odot D_n(j)
\end{equation}
In the question-centric path, attention is performed column-wise on the matching matrix. Every word $S(i)$ in question-answer gets a summarization read $R_n^Q(i)$ of related information in document:
\begin{equation}
\small
	\alpha_n(i,j) = softmax(M_n(i,1),...,M_n(i,L_D))(j)
\end{equation}
\begin{equation}
\small
	R_n^Q(i) = \sum_{j=1}^{L_D}\alpha_n(i,j)D_n(j)
\end{equation}
In the document-centric path, row-wise attention is performed to read related information in the question. Next, cross-document attention is performed on attention reads of all the documents ($\oplus$ represents vector concatenation):
\begin{equation}
    \fontsize{8}{9}\selectfont
	M'_{mn}(i,j) = (D_m(i)\oplus R_m^D(i)) \odot (D_n(j)\oplus R_n^D(j))
\end{equation}
\begin{equation}
    \fontsize{8}{9}\selectfont
\begin{aligned}
	\beta_{mn}(i,j) = softmax&(M'_{m1}(i,1),...M'_{m1}(i,L_D),...\\
	&M'_{mN}(i,1),...,M'_{mN}(i,L_D))_n(j)
\end{aligned}
\end{equation}
\begin{equation}
    \fontsize{8}{9}\selectfont
	R_m^{'D}(i) = \sum_{n=1}^{N}\sum_{j=1}^{L_D}\beta_{mn}(i,j)(D_n(j)\oplus R_n^D(j))
\end{equation}
The cross-document attention extracts relevant information from other documents based on the current document, which enables integration of information from multiple documents.

\begin{figure}[h]
  \centering
    \includegraphics[width=\columnwidth]{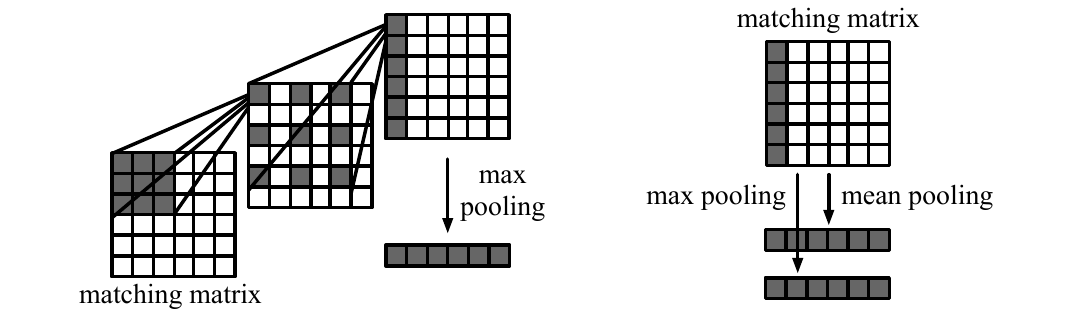}
  \caption{Extract matching features. Left: CNN extractor, right: pooling extractor}
  \label{fig:mf}
\end{figure}

We introduce matching features $F$ as a complement to attention reads.
Softmax in attention destroys absolute matching strength, e.g. ``clinic" should matches ``hospital" better than ``physician", regardless of the accompanying words in the document. We thus extract matching features directly from the matching matrix, using a two-layer convolutional network on $M$ (see Figure \ref{fig:mf} left). Max-pooling is performed between and after convolution layers. The second layer used dilated convolution to keep resolution at the word-level.

The convolution captures patterns in the matching matrix at a significant cost of computation. We also experimented a simpler design with similar performance gain: extracting matching feature by max-pooling and mean-pooling rows and columns in $M$ (Figure \ref{fig:mf} right).

\subsubsection{Reasoning layer} The reasoning layer takes attention reads from the Q-centric and the D-centric path as well as matching features.  It then uses a bi-directional LSTM layer to reason on the question/document level. A gating layer is applied before LSTM to decide whether a word should contribute to reasoning. The gate value is computed from contextual embedding, and is multiplied to all input features.

The outputs of LSTM represent the support of the document to the statement, which is summarized into a single vector by a max-pooling over the sequence.

\subsubsection{Integration \& decision layer} To integrate support from multiple documents, feature vectors first go through a gating layer similar to that in the reasoning layer. Gating decides relevant documents and suppresses irrelevant ones. Next, max-pooling and mean-pooling are performed to further summarize the support of all the documents. The intuition of using two different pooling together is based on the assumption that best candidates should have \textit{better} as well as \textit{more} support, respectively. A multi-layer feed-forward network is used to predict a scalar score. The candidate answer having the largest score value is chosen as the best answer predicted by the model.

\subsection{Interpretable Reasoning}
Neural network models are sometimes regarded as black-box optimization for difficulty interpreting model's behavior. Interpretability is recognized as invaluable for analyzing and improving the model. Here we present a series of novel techniques to improve the interpretability of our SeaReader, which can also be applied to general neural network based NLP models.
\subsubsection{Gating with importance penalty}
The value of a scalar gate can usually be interpreted as the importance of the gated information. However, putting a gate somewhere in the model does not always make its output meaningful. The gate can be passed-through if it does not help optimization. We introduce a regularization term in the objective function to use a gate for inspection purposes:
\begin{equation}
\small
	loss = loss_{task} + c\cdot max(\frac{1}{L}\sum_{i}^{L}g(i)-t,0)
\end{equation}
The added term restricts the average gate value to be below a threshold of $t$ (set to 0.7 in our experiments).  The model must now learn to suppress unimportant information by giving lower gate values to those vectors.

\subsubsection{Noisy gating}
It is sometimes desirable for a model to rely more on strong evidence than weak evidence because the latter contains more noise and makes the model more prone to overfitting. Gating helps to differentiate strong evidence from weaker evidence, and the effect can be reinforced by adding Gaussian noise after gating:
\begin{equation}
\small
	X_{out} = gate(X_{in})X_{in} + \sigma(0, s)
\end{equation}
A high gate value helps a strong evidence stand out from noise. Weaker evidence become harder to utilize in the presence of noise. We found strong  evidence getting higher gate values and otherwise for weak evidences than without the added noise. This helps interpreting gate values and can improve generalization performance.

\subsubsection{$\bm{L}_{21}$ regularized embedding learning}
Word embeddings often have the largest number of parameters in a NLP model. When there is little labeled information for a complex task, it is difficult to learn embeddings well. In our experiments on MedQA, learning word embeddings directly suffers from severe overfit, even with pre-trained embedding as initialization. Fixed word embedding results in decent performance, but leads to an underfit model. To address this problem, we introduce a delta embedding term and adapt $L_{21}$ regularization that is often used in structural optimization \cite{zhou2011malsar,kong2011robust}: 
\begin{equation}
\small
	\bm{w} = \bm{w}_{skip\hbox{-}gram} + \bm{w}_\Delta
\end{equation}
\begin{equation}
\small
	loss = loss_{task} + c\sum_{i=1}^n(\sum_{j=1}^dw_{\Delta ij}^2)^\frac{1}{2}
\end{equation}
Intuitively, the model learns to refine skip-gram embeddings for supervised task while avoiding overfit by modifying only a few words for as little as possible. This does not only improve performance, but also lets us interpret semantics learned from the task, by inspecting \textit{which} words are adjusted and \textit{how} they are adjusted.

\section{Experiments}

\subsection{Experimental Setup}
\subsubsection{Word embedding}
Word embedding is trained on a corpus combining all text from the problems in the training set and the document collection using skip-gram \cite{mikolov2013efficient}. The dimension is set to 200. Unseen words during testing are mapped to a zero vector.
\subsubsection{Model settings}
All documents are truncated to no more than 100 words before processed by SeaReader. Although leading to a minor performance drop, this greatly accelerates the experiments by saving training time on the GPU. In most experiments, we retain 10 documents for each candidate answer, a total of 50 documents per problem. 

Bidirectional LSTMs in the context layer and the reasoning layer all have a dimension of 128. Parameters are shared between LSTMs processing question and documents in the context layer. A single layer feed-forward network is used in the decision layer because more layers did not further improve performance.

\subsubsection{Training} We put a softmax layer on top of candidate score predictions and use cross entropy w.r.t. the groundtruth choice as objective function. Our model is implemented using Tensorflow \cite{abadi2016tensorflow}. Adam optimizer is used with $\epsilon=10^{-6}$ to stabilize training. Exponential decay of learning rate and dropout rate of 0.2 was used to reduce overfit. We used a batch size of 15, which already contains 750 documents per batch and is the maximum allowed to train on a single GPU.

\subsection{Baseline Approaches}
To compare the performance of our SeaReader with existing reading comprehension models, we selected a few models with different considerations and adapted them to our MedQA task:
\begin{itemize}
	\item \textbf{Iterative Attention} \cite{sordoni2016iterative}: This is one of the rare models not extensively tailored for cloze (or span) style tasks. It uses attention to read from question and document alternatively, and the read is performed iteratively with the final state used to make prediction. The model is directly applicable to our MedQA task.
	\item \textbf{Neural Reasoner} \cite{peng2015towards}: A framework for reasoning over natural language sentences, using a deep stacked architecture. It can extract complex relations from multiple facts to answer questions. The model is a natural fit to our task and is straightforward to apply.
	\item \textbf{R-NET} \cite{rnet}: A recent reading comprehension model, achieving state-of-the-art single model result on the challenging SQuAD dataset\footnote{https://rajpurkar.github.io/SQuAD-explorer/, by the time of writing this paper}. It stacks a question-to-document attention layer and a document-to-document attention layer. We replaced the final prediction layer with a pooling layer to generate a scalar score.
\end{itemize}

\subsection{Experimental Results}

\begin{table}[h]
  \small
  \centering
  \caption{Results (accuracy) of SeaReader and other approaches on MedQA task}
  \label{tab:eval}
  \begin{tabular}{lcc}
    \toprule
     & Valid set & Test set  \\
    \midrule
    Iterative Attention & 60.7 & 59.3\\
    Neural Reasoner & 54.8 & 53.0\\
    R-NET & 65.2 & 64.5\\
    SeaReader & \textbf{73.9} & \textbf{73.6}\\
    SeaReader (ensemble) & \textbf{75.8} & \textbf{75.3}\\
    \midrule
    Human passing score\footnotemark & \multicolumn{2}{c}{60.0 (360/600)}\\
    \bottomrule
  \end{tabular}
\end{table}
\footnotetext{of the year 2016}

\subsubsection{Quantitative Results} We evaluated model performance by  accuracy of choosing the best candidate. The results are shown in Table \ref{tab:eval}. Our SeaReader clearly outperforms baseline models by a large margin. We also include the human passing score as a reference. As MedQA is not a commonsense question answering task, human performance relies heavily on individual skill and knowledge. A passing score is the minimum score required to get certified as a Licensed Physician and should reflect a decent expertise in medicine.

\begin{figure}[h]
\centering
\pgfplotsset{width=\columnwidth,compat=1.5,
/pgfplots/ybar legend/.style={
    /pgfplots/legend image code/.code={%
       \draw[##1,/tikz/.cd,yshift=-0.25em]
        (0cm,0cm) rectangle (8pt,0.5em);},font=\scriptsize
   }}
\begin{tikzpicture} \begin{axis}[
    ybar,
	ymin=0.2,
	ymax=1.05,
    height=5.5cm,
    width=8cm,
    bar width=7pt,
    enlarge x limits=0.2,
    legend cell align={left},
    legend style={row sep=-2pt, at={(0.985,0.98)},anchor=north east, inner sep=1pt},
    ylabel={Accuracy},
    symbolic x coords={A1,B1,A2,A3/A4},
    xtick=data,
]
\addplot[fill=magenta!80!red] coordinates {(A1,0.558) (B1,0.531) (A2,0.649) (A3/A4,0.657)};
\addplot[fill=green!90!red] coordinates {(A1,0.530) (B1,0.479) (A2,0.563) (A3/A4,0.559)};
\addplot[fill=blue!70!white] coordinates {(A1,0.671) (B1,0.617) (A2,0.663) (A3/A4,0.618)};
\addplot[fill=violet] coordinates {(A1,0.754) (B1,0.738) (A2,0.737) (A3/A4,0.707)};
    \legend{Iterative Attention,Neural Reasoner,R-NET,SeaReader}
\end{axis}
\end{tikzpicture}
\caption{Accuracy by problem category}
\label{fig:acc_cat}
\end{figure}
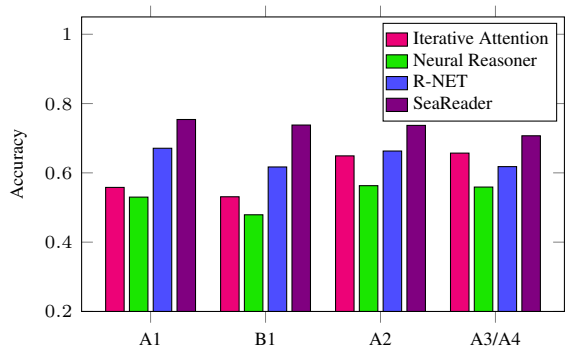

Test performance is broken down by problem category in Figure \ref{fig:acc_cat}. Models with extensive word-level attention (SeaReader and R-NET) win on statement type problems (A1, B1), indicating better ability at capturing details and fine-grained reasoning. The dual-path attention architecture in SeaReader gives a major boost in performance,  especially at solving more complex problems (B1, A3/A4, where information is mixed/shared among problems), showing the effectiveness of multi-perspective information extraction and and integration of information in multiple documents.
 
\begin{table}[h]
  \fontsize{8}{9}\selectfont
  \centering
  \caption{Test performance with different number of documents given per candidate answer}
  \label{tab:num_doc}
  \vskip 0.02 in
  \begin{tabularx}{\columnwidth}{Xcccc}
    \toprule
     Number of documents & top-1 & top-5 & top-10 & top-20\\
    \midrule
    SeaReader accuracy & 57.8 & 71.7 & 73.6 & 74.4\\
    Relevant document ratio & 0.90 & 0.54 & 0.46 & 0.29\\
    \bottomrule
  \end{tabularx}
\end{table}

Test performance clearly increases as more documents is given to SeaReader as input (see Table \ref{tab:num_doc}). To discover the general relevancy of documents returned by our retrieval system, we hand-labeled the relevancy of 1000 retrieved documents for 100 problems. The ratio of documents containing relevant information in the top-N documents is given in Table \ref{tab:num_doc}. We notice that there is still performance gain using as many as 20 documents per candidate answer, while the ratio of relevant documents is already low. This illustrates SeaReader's ability to discern useful information from large-scale text and integrate them in reasoning and decision-making.

\subsubsection{Discussion}
In the extensive architecture search developing SeaReader on MedQA task, we discovered that: 1) MedQA is a complex task with weak labels, and the best model design in such a scenario is a modular architecture without excessive depth. Our SeaReader is not only the best performing but also the fastest to converge in training. Multi-task learning is another important direction in such a scenario. 2) Extensive use of attention compensates for lack of depth, and intra-document and inter-document attention helps to model information flow at different levels, which is crucial for utilizing large-scale text inputs.

\begin{figure}[h]
  \centering
    \includegraphics[width=\columnwidth]{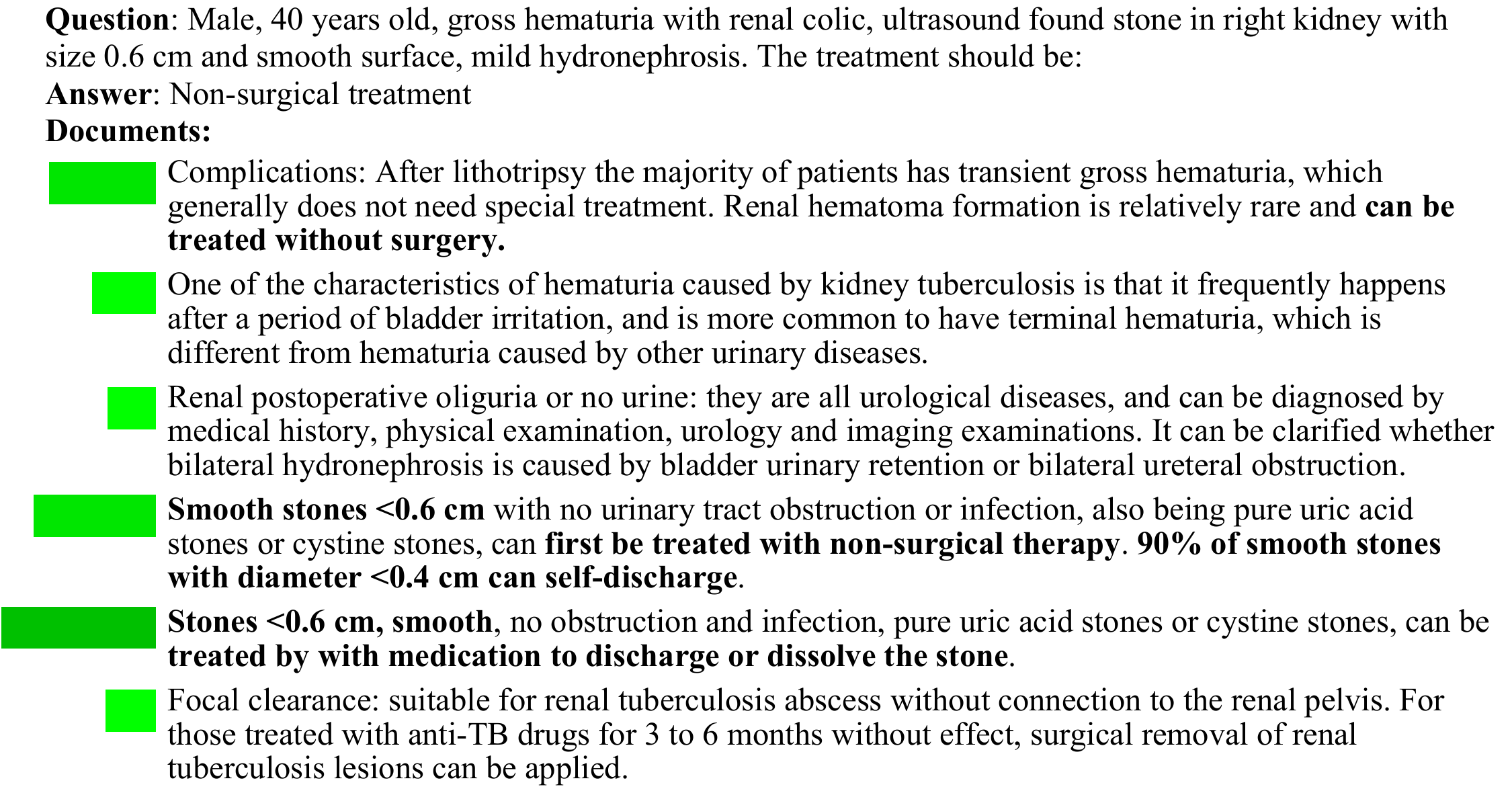}
   \caption{Document-level gating visualization. Bar length is proportional to gate value. Documents are truncated for better visualization.}
  \label{fig:doc}
\end{figure}

\subsubsection{Interpret Reasoning}
The modular design combined with Interpretable Reasoning extensions enabled convenient interpretation of the decision-making process in SeaReader. Using noisy gating on document gating layer, the contribution of each document is clearly represented by gate value (see Figure \ref{fig:doc} for an example). Note that the most contributing document is a semantically relevant one, without sharing many common words with question and answer.

\begin{figure}[h]
\begin{minipage}{.48\columnwidth}
  \centering
  \includegraphics[width=\linewidth]{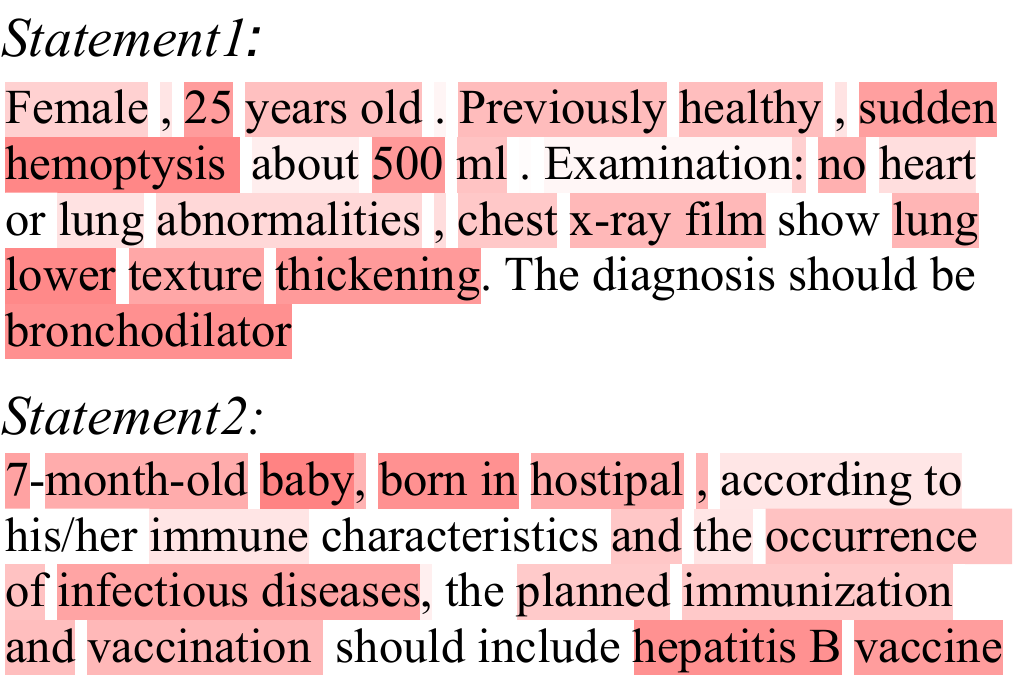}
  \caption{Word-level gating visualization. A darker color indicates larger gate value}
  \label{fig:word}
\end{minipage}
\begin{minipage}{.1\columnwidth}
\end{minipage}
\begin{minipage}{.48\columnwidth}
  \centering
  \includegraphics[width=\linewidth]{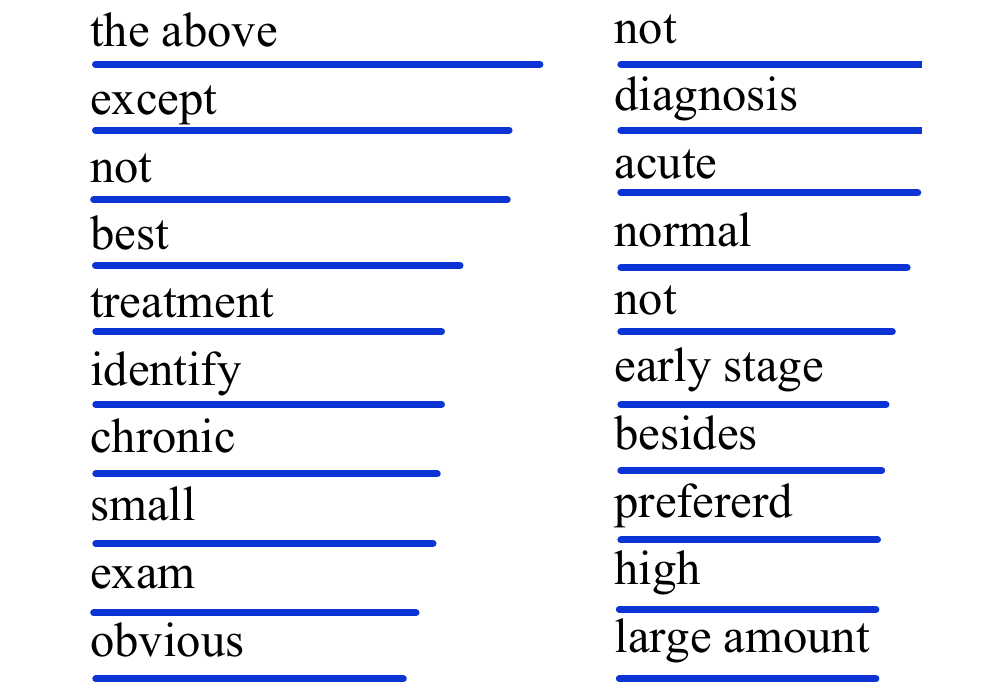}
  \caption{Top-20 words learned with delta embedding\footnotemark. Line length is proportional to norm of $w_\Delta$}
  \label{fig:reg_emb}
\end{minipage}
\end{figure}

The word gating layer gives relative importance of each word when regulated with importance penalty (see Figure \ref{fig:word}). Same words can get different importance by the context they appear in.

Regulating embedding with $L_{21}$ loss allowed us to discover the ``mostly learned", or most relevant words for the task. Figure \ref{fig:reg_emb} shows that the most significant words for MedQA  are words representing logical relations plus some commonly used terms in diagnosis.

Attention weights in SeaReader provide direct illustration about matching and relevance between words in contexts. One is able to examine the extracted information from question, documents and across documents. We omit detailed example here for brevity of the paper. 
\footnotetext{these are translations of corresponding Chinese words}
\section{Conclusion}
In this work, we introduced a question-answering task MedQA on clinical medicine and our reading comprehension model SeaReader. The unique challenge of MedQA is to solve advanced exam problems with large-scale text reading. Our solution SeaReader uses a dual-path attention architecture to integrate information from multiple text documents. Experiments validated its effectiveness especially at solving complex problems and utilizing information. We also propose techniques to help interpret question-answering models, which help to explain the reasoning process of SeaReader. We hope this work contributes to real-world question answering and the application of deep-learning approaches in medicine.

\bibliography{unt}
\bibliographystyle{aaai}

\end{document}